\documentclass{article}

\usepackage{microtype}
\usepackage{graphicx}
\usepackage{subcaption}
\usepackage{booktabs} 

\usepackage{hyperref}

\usepackage[preprint]{icml2026}

\usepackage{amsmath}
\usepackage{amssymb}
\usepackage{mathtools}
\usepackage{amsthm}

\usepackage[capitalize,noabbrev]{cleveref}

\theoremstyle{plain}

\theoremstyle{definition}

\theoremstyle{remark}

\usepackage[textsize=tiny]{todonotes}

\icmltitlerunning{Are Your Reasoning Models Reasoning or Guessing? A Mechanistic Analysis of Hierarchical Reasoning Models}

\begin{document}

\twocolumn[
  \icmltitle{Are Your Reasoning Models Reasoning or Guessing? \\ 
  A Mechanistic Analysis of Hierarchical Reasoning Models}



  \icmlsetsymbol{equal}{*}

  \begin{icmlauthorlist}
    \icmlauthor{Zirui Ren}{qizhi,tsinghua}
    \icmlauthor{Ziming Liu}{qizhi,tsinghua2}
  \end{icmlauthorlist}

  \icmlaffiliation{qizhi}{Shanghai Qi Zhi Institute, Shanghai, China}
  \icmlaffiliation{tsinghua}{Department of Physics, Tsinghua University, Beijing, China}
  \icmlaffiliation{tsinghua2}{College of AI, Tsinghua University, Beijing, China}

  \icmlcorrespondingauthor{Zirui Ren}{rzr23@mails.tsinghua.edu.cn}
  \icmlcorrespondingauthor{Ziming Liu}{lzmsldmjxm@gmail.com}

  \icmlkeywords{latent-space reasoning, mechanistic interpretability, data augmentation, inference-time scaling}

  \vskip 0.3in
]



\printAffiliationsAndNotice{}  

\begin{abstract}
Hierarchical reasoning model (HRM) achieves extraordinary performance on various reasoning tasks, significantly outperforming large language model-based reasoners. To understand the strengths and potential failure modes of HRM, we conduct a mechanistic study on its reasoning patterns and find three surprising facts: (a) \textbf{Failure of extremely simple puzzles}, e.g., HRM can fail on \textit{a puzzle with only one unknown cell}. We attribute this failure to \textit{the violation of the fixed point property}, a fundamental assumption of HRM. (b) \textbf{``Grokking'' dynamics in reasoning steps}, i.e., the answer is not improved uniformly, but instead there is a \textit{critical reasoning step} that suddenly makes the answer correct; (c) \textbf{Existence of multiple fixed points}. HRM ``guesses'' the first fixed point, which could be incorrect, and gets trapped there for a while or forever. All facts imply that \textbf{HRM appears to be ``guessing'' instead of ``reasoning''}. Leveraging this ``guessing'' picture, we propose three strategies to scale HRM's guesses: data augmentation  (scaling the quality of guesses), input perturbation (scaling the number of guesses by leveraging inference randomness), and model bootstrapping (scaling the number of guesses by leveraging training randomness). On the practical side, by combining all methods, we develop \textbf{Augmented HRM}, \textbf{boosting accuracy on \textit{Sudoku-Extreme} from 55.0\% to 96.9\%}. On the scientific side, our analysis provides new insights into how reasoning models ``reason".

\end{abstract}

\section{Introduction}
\label{introduction}
Current large language models (LLMs) are based on the transformer architecture~\cite{vaswani2023attentionneed}. Despite their success, they still struggle in reasoning-intensive tasks. For example,~\citet{wang2025hierarchicalreasoningmodel} showed that even the best language models achieve 0\% success rate in solving hard Sudokus and mazes. In an effort to build models competent for these reasoning tasks, which generally require systematic System-2 thinking for humans, massive work has been done to prolong the reasoning process of LLMs, i.e., the intermediate outputs before reaching the final answer. It either relies on chain-of-thought (CoT) prompting~\cite{wei2023chainofthoughtpromptingelicitsreasoning, chen2025survey} or fine-tuning with reinforcement learning~\cite{guo2025deepseek,wen2025reinforcement,yue2025does}, both of which perform reasoning at the token level, limiting the potential of deep networks~\cite{turpin2023languagemodelsdontsay, helwe2021reasoning}.

Alternatively, latent-space reasoning models~\cite{hao2025traininglargelanguagemodels, wang2025hierarchicalreasoningmodel} rise as a new paradigm of reasoning depth scaling. Among them, the hierarchical reasoning model (HRM)~\cite{wang2025hierarchicalreasoningmodel} achieves extraordinary accuracy on various reasoning-intensive tasks, outperforming LLM reasoners by a significant margin.

To understand the secret sauce of the success of HRM and reveal its potential failure modes, we closely inspect the reasoning patterns of HRM, mainly focusing on the \textit{Sudoku-Extreme} dataset~\cite{wang2025hierarchicalreasoningmodel}. To our surprise, we identify three counterintuitive facts:

\begin{figure*}[t]
\begin{center}
\centerline{\includegraphics[width=\textwidth]{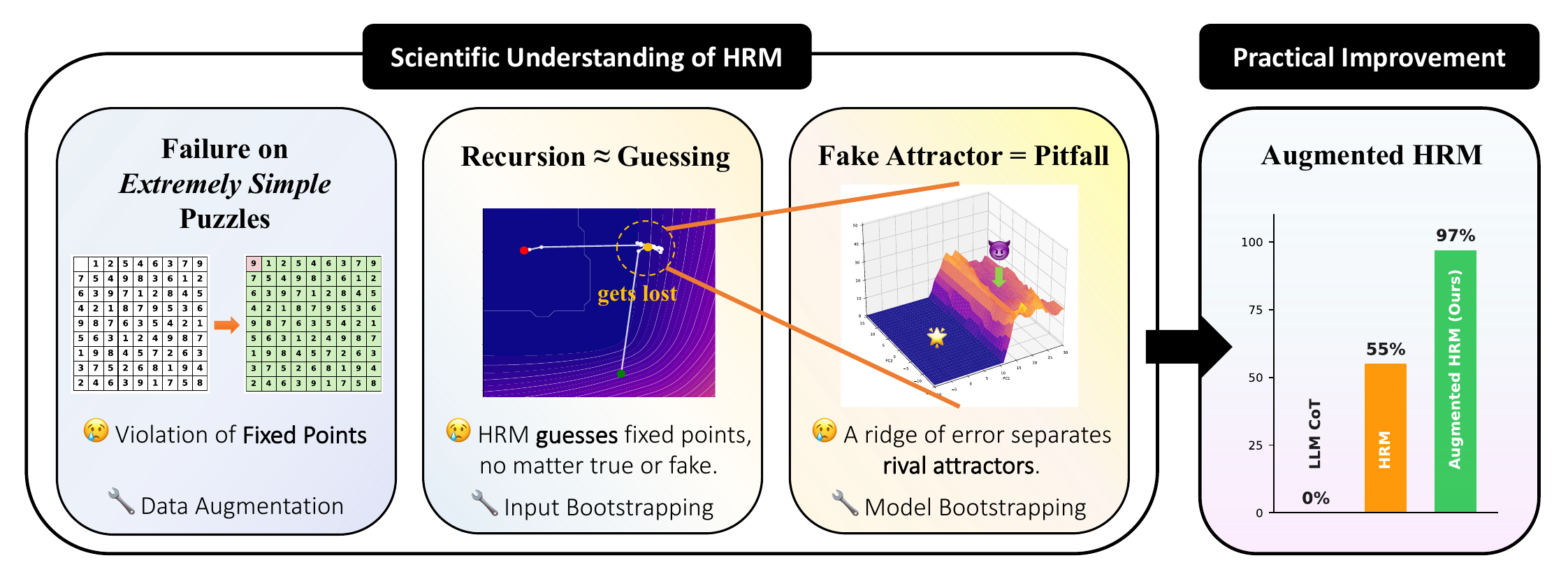}}
\caption{A lone unknown cell exposes the fixed-point violation. HRM secretly guesses fixed points, no matter they are true or not. Multiple fixed points exist in the latent space; escaping them via data augmentation, input and model bootstrapping boosts accuracy from 55.0\% to 96.9\%.}
\label{visual_abstract}
\end{center}
\vskip -0.1in
\end{figure*}

\begin{itemize}
    \item 
    \textbf{Failure on \emph{extremely simple} puzzles, due to violation of fixed points (Section 3)}: HRM could fail on \textit{a puzzle with only one unknown cell}, due to a theory-practice mismatch. HRM theory assumes the fixed point property, i.e., the ability to maintain stability after finding the solution; however, we find that this property breaks down in practice. 
    Luckily, we find that a simple fix suffices -- data augmentation. 

    
    \item 
    \textbf{``Grokking'' dynamics in recursion, due to HRM ``guessing" instead of ``reasoning" (Section 4)}: When approaching a puzzle, HRM does not incrementally refine the answer at each recursive step. Instead, it typically gets completely perplexed (error remains high and flat for many steps), and then ``groks'' (error drops to zero in one step). We hypothesize that the recursion (outermost loop) of HRM serves as a way of scaling ``guessing" attempts for a plausible latent state, challenging the common belief that recursive reasoning boosts performance by incremental refinement.

    \item \textbf{Reasoning ``gets lost" in the latent space, due to multi-stability of the reasoning landscape (Section 5)}: Closely inspecting the reasoning trajectory in the latent space, we are able to classify reasoning modes of HRM, among which the most interesting failure mode is when it ``gets lost", i.e., lingering around some misleading attractive point.
    We show that these false attractors can be interpreted as local optima of a heuristic error metric measuring the number of conflicts. This trap discourages HRM from further exploring the latent space, postponing or precluding the encounter of the ``true'' fixed point. It turns out to be the central factor that caps HRM at suboptimal accuracy.
\end{itemize}

All insights above imply that HRM appears to be ``guessing" instead of ``reasoning". In order to get better performance, the ``guessing" picture points to a new scaling axis -- guess attempts, in addition to model and data. We thus propose three methods to scale guessing attempts: data augmentation (scaling the quality of guesses), input perturbation (scaling the number of guesses by leveraging inference randomness), and model bootstrapping (scaling the number of guesses by leveraging training randomness). Combining all methods, our \textbf{Augmented HRM} is able to boost the accuracy from 55.0\% to 96.9\% for \textit{Sudoku-Extreme}, surpassing vanilla HRM~\cite{wang2025hierarchicalreasoningmodel} and its existing variants such as the Tiny Recursive Model~\cite{jolicoeurmartineau2025morerecursivereasoningtiny}. Scientifically, our findings provide new insights into how reasoning models could ``reason". Our experimental code is available at \url{https://github.com/renrua52/hrm-mechanistic-analysis}.

\section{Background on HRM}

HRM is a recursive latent-space model in the sense that in each forward pass, the inputs and hidden states are recurrently passed through the same module several times. Each call of this module is named a \emph{segment}, and the recursive loop of segments is named the \emph{outer loop}.\footnote{The original HRM architecture uses separate H-module and L-module within one segment. However, ablation studies have shown that this structure is not the core factor of superior performance~\cite{ge2025hierarchicalreasoningmodelsperspectives, jolicoeurmartineau2025morerecursivereasoningtiny}. It is also not relevant to our analysis of reasoning trajectories in following sections. Regarding these facts, in this paper we abstract away this inner structure of segments. This is merely a simplification of notation and does not alter the model architecture; see Appendix~\ref{simplification} for details of this simplification.}

\subsection{Forward Pass}
\label{forward_pass}

We formalize the (simplified) forward pass of HRM as follows. The input sequence is mapped to its embedding $\tilde{x}$ by the input network $f_I$: 
\begin{equation}
    \tilde{x}=f_I(x;\theta_I)
\end{equation}
In this paper, we mainly focus on the \textit{Sudoku-Extreme} dataset. The input samples $x$'s are sudoku puzzles, formatted as $9\times9$ sequences containing integer tokens ranging from $1$ to $9$, together with a special \texttt{<blank>} token representing masked cells.

As a latent-space model, HRM maintains a latent state $z^i$, deterministically initialized as $z^0$. The HRM segment $\mathcal{F}$ takes the input embedding and the current latent state as input, and computes the next latent state:
\begin{equation}
\label{forward_z}
    z^{i+1}=\mathcal{F}(z^i,\tilde{x};\theta)
\end{equation}
After all $M$ segments, the prediction vector is extracted from the terminal latent state:
\begin{equation}
    \hat{y}=f_O(z^M; \theta_O)
\end{equation}

In practice, when HRM has already reached a plausible solution, the remaining segments are essentially redundant. Accordingly, an adaptive computation time (ACT) mechanism~\cite{graves2017adaptivecomputationtimerecurrent} is introduced to decide whether to halt the computation after each segment.

\subsection{Deep Supervision \& One-step Gradient}

\subsubsection{Reasoning Depth Scaling}

The core training technique to achieve reasoning depth scaling is \emph{deep supervision}~\cite{wang2025hierarchicalreasoningmodel}. Each forward pass corresponds to only one ground truth label, while the number of segments can be arbitrarily scaled. This mismatch makes the loss signal sparse compared to reasoning depth.

Deep supervision addresses this issue by computing the loss for the latent state $z^i$ and the associated output $\hat{y}^i=f_O(z^i,\tilde{x};\theta_O)$ of \emph{each} segment. Formally, 
\begin{equation}
    L^i = l(\hat y^i, y)
\end{equation}
where $l$ is the loss function. However, the computation required for a full back propagation through time (BPTT)~\cite{Rumelhart1986LearningIR, werbos1990bptt} of all these segment losses scales at $\mathrm{\Theta}(T)$, and deep supervision at each segment would cost $\mathrm{\Theta}(T^2)$ in total. 

To overcome this, a dual technique called \emph{one-step gradient}~\cite{wang2025hierarchicalreasoningmodel} is used: during optimization, the gradient of $L^i$ is only computed with respect to the $i$-th segment. In other words, $z^i$ is detached from the computational graph each time it is updated via \cref{forward_z}. It guarantees that training costs increase at the same rate as reasoning depth.

\subsubsection{The Role of Fixed Point Property}
\label{fixed_point}

The one-step gradient alters the standard approach of training RNN-like models, and thus needs further theoretical grounding. ~\citet{wang2025hierarchicalreasoningmodel} gave their justification, based on a few assumptions. One is that $\mathcal{F}$ is continuously differentiable. More importantly, they made the plausible presumption that if HRM finds $z^*$ whose associated output is correct, it does not make further updates to the latent state. Thus, $z^*$ should be the fixed point of $\mathcal{F}$, satisfying
\begin{equation}
\label{fp_eq}
    z^*=\mathcal{F}(z^*, \tilde{x};\theta)
\end{equation}
Differentiating \cref{fp_eq} gives ($J_{\mathcal{F}}\equiv \frac{\partial \mathcal{F}}{\partial z}$):
\begin{equation}
\label{grad_approx}
    \frac{\partial z^*}{\partial \theta} = \left(I-\left.J_{\mathcal{F}}\right\rvert_{z^*}\right)^{-1} \left.\frac{\partial \mathcal{F}}{\partial \theta}\right\rvert_{z^*} \approx \left.\frac{\partial \mathcal{F}}{\partial \theta}\right\rvert_{z^*}
\end{equation}
which shows that as long as $I-\left.J_{\mathcal{F}}\right\rvert_{z^*} \approx I$ (a common approximation for implicit models), we can substitute the full BPTT gradient with the one-step gradient~\cite{geng2022trainingimplicitmodels}.

This succinct argument clearly relies strongly on the fixed point assumption. However, despite being intuitive and plausible, this assumption does not hold trivially. Our experiments demonstrate that this violation has profound consequences, to be discussed in \cref{fp_hrm}.

\subsection{Evaluation of HRM}

The target benchmark is \textit{Sudoku-Extreme}~\cite{wang2025hierarchicalreasoningmodel}, consisting of extremely difficult Sudoku puzzles. HRM is trained on the augmented versions of 1000 samples from the training data set, which are of the same difficulty level as the test set. HRM achieves a remarkable 55\% accuracy on \textit{Sudoku-Extreme}, demonstrating strong generalization abilities, while o3-mini-high, Claude 3.7 8K and Deepseek R1 all fail completely on this extremely complex task, achieving 0.0\% success rate~\cite{wang2025hierarchicalreasoningmodel}.

\section{Failure on Extremely Simple Puzzles: Violation of Fixed Points}
\label{fp_hrm}

The fixed point property, which states that the latent state is no longer updated after the true answer is found, is the fundamental assumption of HRM. It helps establish \cref{grad_approx}, which justifies the one-step gradient approximation, in turn enabling reasoning depth scaling. Due to its simplicity and importance, it is a natural question whether HRM indeed demonstrates the fixed point property.

To our surprise, when dealing with \emph{extremely simple} puzzles (e.g. with only one cell to fill), HRM does not retain the correct solutions after solving these puzzles in very early segments. In extreme cases, this instability manifests itself even earlier, resulting in a complete failure on such samples. This phenomenon is uncharacteristic of such a strong model, and causes tangible loss to performance.

In this section, we discuss this counterintuitive phenomenon of fixed point violation. We also present our explanation and a straightforward fix.

\subsection{Violation of Fixed Point Assumption}
\label{violation}

\begin{figure*}[t]
\begin{center}
\centerline{\includegraphics[width=0.8\textwidth]{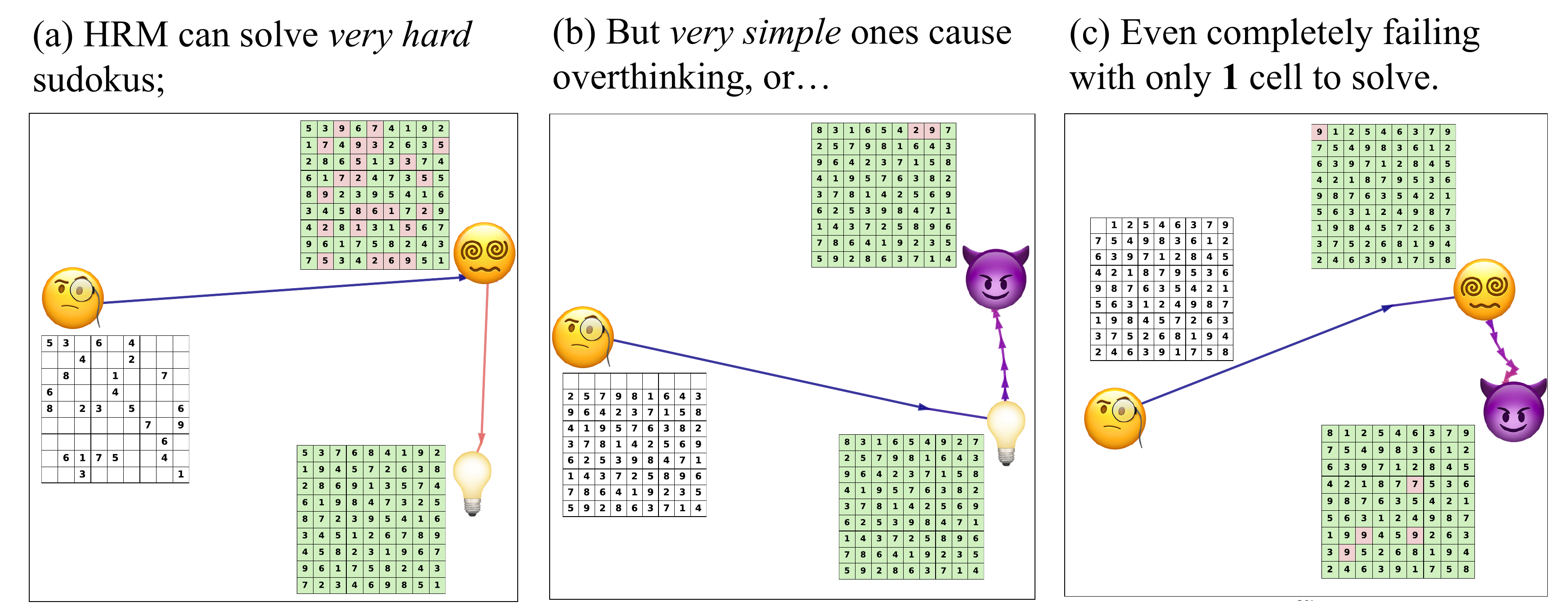}}
\caption{Reasoning trajectories in latent space (projected onto first two principal components) for different sudoku puzzles, and associated answers (red cell: wrong, green cell: correct). (a) For a difficult sudoku from the validation set of \textit{Sudoku-Extreme}, the answer is correct; (b) For a simple puzzle with only one row masked, the answer is correct for the first two segments, but the model continues updating to wrong answers, violating the fixed point assumption; (c) Complete failure on an \emph{extremely simple} puzzle with only one masked token.}
\label{fp_figure}
\end{center}
\vskip -0.1in
\end{figure*}

Indeed, for the test samples from \textit{Sudoku-Extreme}, if HRM was able to reach the solution at any segment, the remaining segments do not alter the output anymore (\cref{fp_figure}.a). Nevertheless, this single observation does not prove that HRM has learned the fixed point property, since these samples are as difficult as the ones HRM has been trained on.

Fixed points emerge at the final stage of HRM reasoning, where the model is very close to the solution. Thus we present a few extremely simple Sudoku puzzles to the same model to probe this property. These puzzles are constructed by masking out only one row, or even only one cell, from a complete sudoku. Shockingly, HRM frequently corrupts its answer by making unnecessary updates to its latent state, even after having reached the correct answer at very early segments (\cref{fp_figure}.b). In fact, HRM is only able to maintain stability on such puzzles about 75\% of the time.

In a slightly rarer case, instability shows up earlier, taking over even before HRM is able to find the solution. The consequence is a peculiar phenomenon: HRM sometimes gets a sudoku completely wrong throughout the segments, even if only one token is missing (\cref{fp_figure}.c).\footnote{Importantly, HRM architecture does not enforce the preservation of unmasked input tokens - it merely maps one sequence to another. In extreme cases, unmasked tokens can be corrupted, rendering the segments completely ineffective.}

\subsection{One-step Gradient Postpones Acquisition of Stability}
\label{instability_explanation}

The ability to maintain fixed points, though proposed as a prior assumption, is by no means automatically obtained and has to be acquired from training. Due to one-step gradients, segments are disentangled and all contribute to the same goal (despite their sequential structure) -- solving extremely hard Sudokus in one step. As a result, HRM is not explicitly trained to complete easier Sudokus, which is the key to ensuring stability around fixed points. Internally, HRM may implicitly create easier problems after some training, when a segment reasonably maps a hard problem into a partial solution (which itself can be viewed as an easier problem for later segments). However, this implicit curriculum is ineffective and not well-controlled. Consequently, the acquisition of stability is postponed to the terminal phase of training, which is far from sufficient.

A helpful analogy is to think of diffusion models, which require data of many noise levels. One-step gradient, combined with \textit{Sudoku-Extreme}, means only the noisiest data (extremely hard puzzles) and the clean data (solutions) are seen, and the goal of HRM is to map the noisiest data directly to the clean data, which is known to be very hard (at least non-trivial) for diffusion models.\footnote{Although not impossible, e.g.,~\cite{frans2025one,geng2025mean}.} Inspired by diffusion models, HRM also needs data at diverse ``noise levels'' (in this case, puzzles of different difficulty levels), as we discuss below.




\subsection{Restoring Fixed Points via Data Augmentation}
\label{data_mixing}

In order to restore the fixed point property of HRM for arbitrary inputs, we seek a direct way of enforcing the acquisition of stability. According to the discussion in \cref{instability_explanation}, we are faced with the question: how do we provide HRM with more opportunities to train its stability?

We can explicitly advance such chances by exposing the model to nearly-solved puzzles directly. This inspires a simple data augmentation: for each puzzle in the training set, we make one simplified replicate. We reveal a random portion of the originally hidden tokens in the replicate, obtaining a simpler version of the puzzle, which is still valid as a training sample.

\begin{figure}[ht]
\vskip 0.2in
\begin{center}
\centerline{\includegraphics[width=\columnwidth]{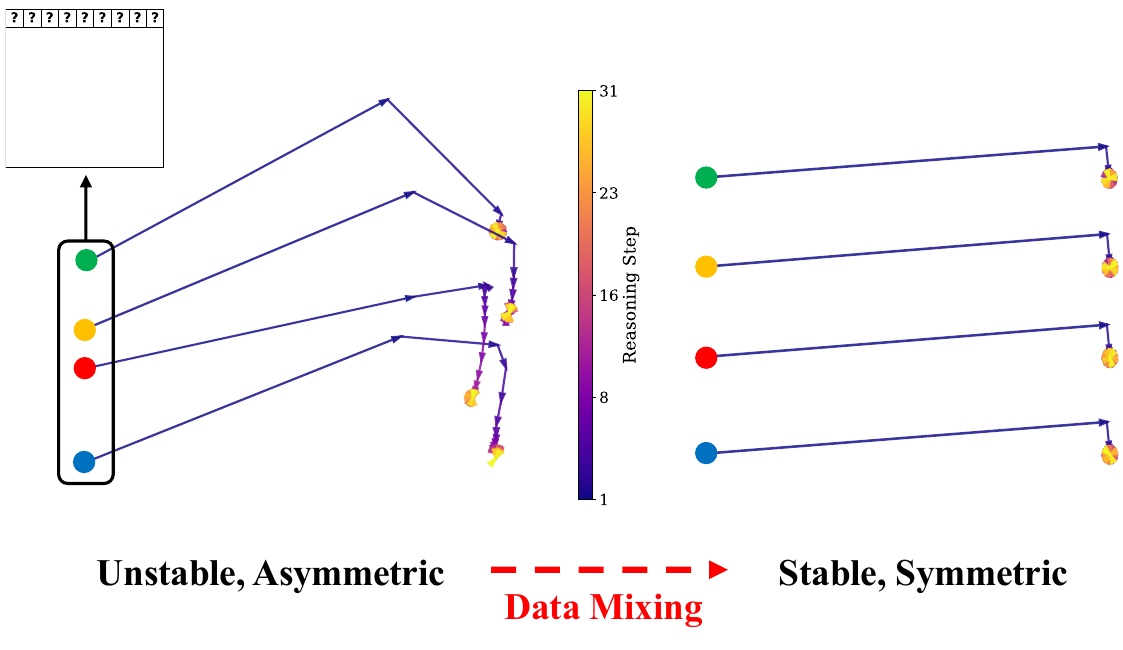}}
\caption{Data mixing restores \textbf{stability} and \textbf{symmetry} of latent reasoning trajectories (projected onto the first two principal components). It eliminates unfavored drifts after getting the correct answer. Furthermore, when dealing with distinct samples simplified via the same format, all latent trajectories now show perfect symmetry.}
\label{refined_hrm_traj}
\end{center}
\end{figure}

By training an HRM model with this augmentation technique, the overall test accuracy is increased from 55.0\% to 59.9\% (see \cref{acc_table}). More importantly, the failure on simple puzzles (discussed in \cref{violation}) is eradicated. Similarly, the unfavored drift after finding the solution is also eliminated (see \cref{refined_hrm_traj}). We conclude that by mixing samples of different complexity levels into training data, the fixed point property of HRM is restored.

Other nice properties emerged as a by-product of fixed point restoration. For example, for puzzles generated via the same simplification formats (e.g., masking the first row), the latent-state trajectories now become symmetric: although these puzzles are different in tokens, they demonstrate nearly identical reasoning trajectories, suggesting the emergence of more general rules instead of overfitting to subtle details. This has not been the case for vanilla HRM (see \cref{refined_hrm_traj}).

\section{Reasoning Modes of HRM: ``Guessing'' instead of Reasoning}

In this section, we examine the way HRM approaches extremely hard puzzles, specifically by looking at the trajectory of its latent states.

\subsection{Mean-field Analysis: Scaling Laws of Loss Curves}

When presented with the formalization of HRM, or any recursive reasoning model, one would expect that scaling reasoning depth benefits the accuracy of the output monotonically. In other words, intuitively $L^i$ should \textit{gradually} decrease with $i$.

\begin{figure}[ht]
\vskip 0.2in
\begin{center}
\centerline{\includegraphics[width=\columnwidth]{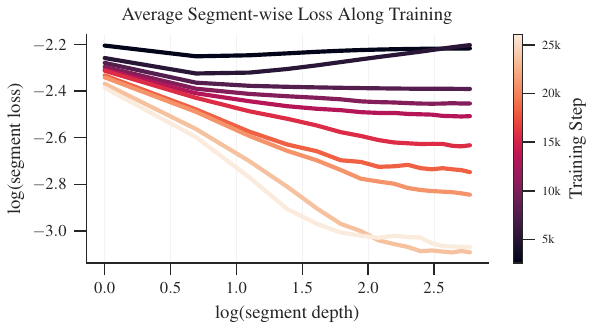}}
\caption{Segment-wise loss reduction improves over training. When averaging across the test samples, more segments lead to smaller losses in a smooth and incremental way, contrasting the ``grokking'' curves for per-sample analysis in \cref{single_loss}. 
}
\label{average_loss}
\end{center}
\end{figure}

We start by pointing out that this is indeed true if we average the loss across all test samples (see \cref{average_loss}). Also, the loss reduction rate steadily increases as training progresses. This observation justifies the repeated utilization of HRM segment: it does improve the latent state iteratively on average.

\subsection{Single-Sample: 
``Grokking'' Along Segments
}
\label{single_grok}

\begin{figure}[hb]
\vskip 0.2in
\begin{center}
\centerline{\includegraphics[width=0.8\columnwidth]{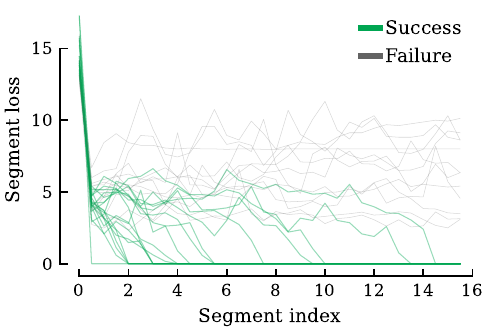}}
\caption{Per-sample analysis shows ``grokking'' dynamics along segments. For success samples, the loss value hovers above some threshold value before suddenly dropping to zero.}
\label{single_loss}
\end{center}
\end{figure}

The surprise comes when we examine how the loss decreases with segments in a single sample. As shown in \cref{single_loss}, the error reduction process is by no means ``gradual''. After being quickly reduced to a relatively low value, the loss enters a lengthy plateau before suddenly dropping to zero. It seems that HRM gets perplexed in most of its segments, hovering around a state with relatively low loss, yet still far from correct. After that, it either remains so for the rest of the time and fails, or suddenly ``groks'' the correct answer in very few segments. This is in sharp contrast to the intuitive picture of gradual loss reduction, where the segments are viewed as individual refining steps. A similar phenomenon is also reported in CoT-based models~\cite{wang2025entropylangletextttthinkrangle} and the quanta model~\cite{michaud2023quantization}.

\subsection{Four Reasoning Modes of HRM}\label{four_modes}

\begin{figure}[ht]
\vskip 0.2in
\begin{center}
\centerline{\includegraphics[width=\columnwidth]{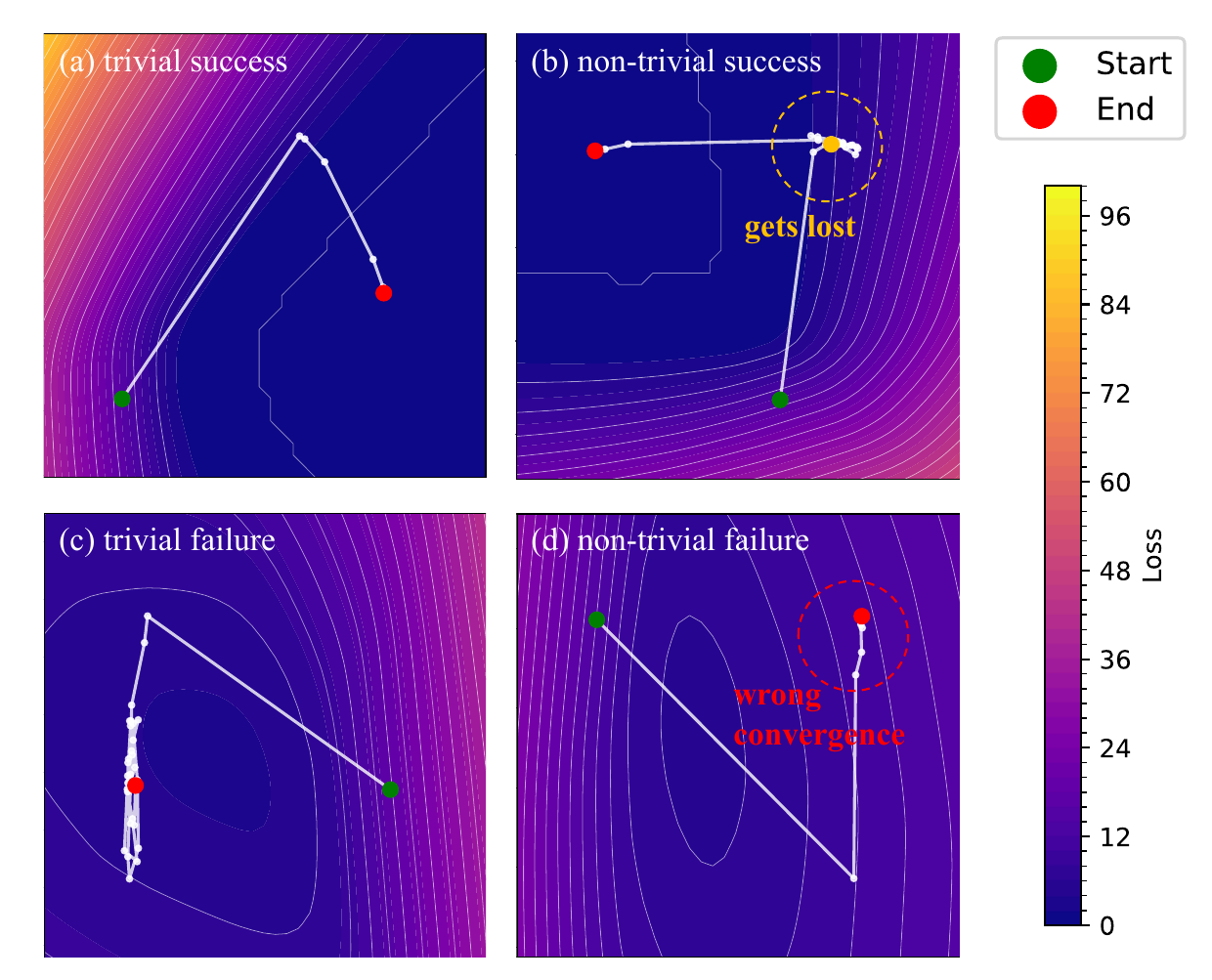}}
\caption{HRM has four reasoning modes on different samples. Trajectories of latent states projected onto the principal plane, with background showing loss values. The four samples demonstrate four modes observed: (a) trivial success; (b) non-trivial success; (c) trivial failure; (d) non-trivial failure.}
\label{fig:four_modes}
\end{center}
\end{figure}

We visually analyze latent reasoning trajectories by projecting them onto the plane defined by their first two principal components. Remarkably, we can classify the reasoning patterns of HRM into the following modes:

\begin{enumerate}
    \item \textbf{Trivial Success} The model finds the solution in the first few segments (\cref{fig:four_modes}.a);
    \item \textbf{Non-trivial Success} The latent state first lingers around some point for quite a few segments, then takes a sudden leap in an orthogonal direction and then immediately finds the solution (\cref{fig:four_modes}.b);
    \item \textbf{Trivial Failure} The latent state wanders about or oscillates in the latent space, encountering nothing special, and error remains high (\cref{fig:four_modes}.c);
    \item \textbf{Non-trivial Failure} The latent state converges to a fixed point. However, this fixed point does \emph{not} correspond to the solution, and the error is in fact still high (\cref{fig:four_modes}.d).
\end{enumerate}

Both trivial success and trivial failure modes are expected: if HRM gets a good latent state early (e.g. initialized close to the solution), it easily reaches a correct fixed point and stays there ever since. On the contrary, if it fails to make any improvement, its inner state wanders about or oscillates in the latent space.

\subsection{Spurious Fixed Points as Misleading Attractors}
\label{misleading_attractors}

Of real interest are the non-trivial cases: they seem to imply that there exists a kind of ``false'' fixed point associated with a wrong output. When entering the neighborhood of such points, it becomes challenging for HRM to leave the region. Such attractors sometimes cause HRM to linger around it for many segments before finally leaping out, leading to the non-trivial success mode; alternatively, if they trap the model forever, the model would converge to a false fixed point (non-trivial failure mode).


\subsection{Escaping the Trap}
\label{perturb}

The aforementioned behavior inspires us to incentivize HRM to escape or avoid such ``traps'' proactively, by adding perturbation to its reasoning trace. The idea is that perturbation increases the chance of encountering a true fixed point, or escaping a false one. 

The implementation of this idea, however, is slightly subtler. Due to the high dimensionality of the latent space, perturbing $z$ in an arbitrary direction typically destructs the coherence of hidden states, leading to unstable behavior.\footnote{We do observe that in some cases, such techniques help HRM solve puzzles that it could not solve before, though. However, no consistent path to success was found.}

To circumvent this difficulty, we point out that there naturally exist indirect ways to exert perturbation in alternative spaces. The input space can be manipulated to create semantically equivalent input sequences with distinct formats. Besides, the adjacent training checkpoints are naturally perturbed versions of the model. Thus, perturbation in the model parameter space is possible, although the space itself is not directly manipulable.

We propose two indirect inference-time scaling techniques to unleash HRM from the trap: \textbf{input perturbation} and \textbf{model bootstrapping}. Both these techniques requires multiple forward passes; each pass reports failure when the ACT mechanism fails to halt within the reasoning depth limit. A majority vote is done among those that successfully halted. Details are elaborated on in the following sections.

Combining these with data augmentation introduced in \cref{data_mixing}, we propose \textbf{Augmented HRM}, reaching the state-of-the-art performance on \textit{Sudoku-Extreme}.

It is worth pointing out that these methods are essentially different from \textit{pass@k}~\cite{chen2021evaluatinglargelanguagemodels} used in LLM evaluation. The next-token prediction of an LLM is \emph{randomized}, which is why multiple passes generate different outputs; these passes, however, are not guaranteed to be intrinsically distinct. Our method leverages perturbation in various working spaces of HRM, creating \emph{intrinsic} diversity in its \emph{deterministic} forward passes.

\begin{table}[t]
\caption{Evaluation results of techniques in the paper: data mixing for fixed point property restoration, input transformation by token relabeling, and model bootstrapping.}
\label{acc_table}
\begin{center}
\begin{small}
\begin{sc}
\begin{tabular}{lcccr}
\toprule
\textbf{Method} & \textbf{Exact Accuracy} \\
\midrule
Baseline & 55.0\% \\ 
+Bootstrap & 64.2\% \\ 
+Relabel & 73.2\% \\ 
+Bootstrap+Relabel & 85.4\% \\ 
+Data Mixing & 59.9\% \\ 
+Data Mixing+Bootstrap & 79.3\% \\ 
+Data Mixing+Relabel & 78.4\% \\ 
+All (\textbf{Augmented HRM}) & \textbf{96.9\%} (SOTA) \\ 
\bottomrule
\end{tabular}
\end{sc}
\end{small}
\end{center}
\vskip -0.1in
\end{table}

\subsubsection{Perturbation in the Input Space}
\label{perturb_input}

It is well known that a sudoku puzzle can be transformed into alternative versions by applying certain types of equivalent transformations. For example, if we have a valid\footnote{By ``valid'' we mean that the sudoku puzzle can be uniquely solved, given the revealed tokens.} puzzle, we can swap the first and the second row, resulting in another valid puzzle. If we knew the solution of this transformed version, an inverse transformation would give the solution of the primal puzzle. Such transformations include relabeling the tokens, swapping bands, swapping rows or columns within a band, and reflecting/rotating the entire puzzle.

Importantly, tokens are treated by HRM as independent features, each assigned with a unique embedding vector by the mapping $f_I$. Consequently, vanilla HRM is intrinsically oblivious to the symmetry under transformations. In other words, HRM views the transformed puzzle as a fresh input. This fact allows us to perturb the input by applying any of the aforementioned transformations to it.

One would expect HRM to perform equally well on these equivalent puzzles, since transformations had been used as data augmentation during training. However, much to our surprise, transforming the input \emph{does} improve the performance. Specifically, we choose the relabeling transformation\footnote{This resonates with the color permuting augmentation used by~\cite{franzen2025productexpertsllmsboosting} when approaching ARC~\cite{chollet2019measureintelligence}}; by creating as few as 9 transformed puzzles via relabeling, we achieve a remarkable 18.2\% improvement in exact accuracy (see \cref{acc_table}).

\subsubsection{Perturbation in the Model Parameter Space}

Another natural space to create perturbed variants is the model parameter space of $\theta$. Altering $\theta$ by force does not work, because such arbitrary perturbation severely imperils model capability. However, the training process offers an indirect way of doing this: the adjacent model checkpoints are approximately equally strong, while the optimizer steps between them diversifies their parameters. Inspired by this, we pick 10 of the checkpoints from the later half of training phase separated by approximately 1000 steps as an ensemble.

As we are testing the ensemble of checkpoints in the same training run, which should be strongly correlated, one might expect that the final strongest checkpoint (receiving the most training) should cover the capability of all its predecessors.
However, we do observe a 9.2\% improvement in accuracy with this model bootstrapping method. 
This strongly contradicts intuition, demonstrating that these mutually dependent models actually differ significantly, in terms of which of the test samples they get right.

\subsection{Reasoning or Guessing?}

The discoveries in \cref{single_grok} have already showed that the segments of HRM do not serve as a cumulative way of refining the output. Further, for the samples on which HRM achieved success, the contribution of each segment are not equal: most of the intermediate steps clustered around the spurious fixed point, making no substantial progress. This shows that HRM is perfectly capable of getting the solution in very few steps, if only already being around the correct latent state. However, it does not really strategize its search in the latent space.

We conclude that HRM does not ``reason'' in the commonsense way of approaching the solution gradually, despite using recursive architecture to mimic human reasoning behavior. If one insists on making an analogy to human intelligence, it resembles ``guessing'' more than ``reasoning''.

Another corroboration of this claim is the fact that the multiple pass techniques in \cref{perturb} benefits performance. Given multiple chances to try approaching the puzzle with different perspectives, HRM is able to achieve much better accuracy; likewise, a human is more likely to be correct with multiple guessing attempts. However, if one is to approach a complex problem through deliberative reasoning, the number of attempts typically matters less.

\section{Spurious Fixed Points}

\cref{misleading_attractors} confirms the existence of spurious fixed points which mislead or trap HRM from exploring more of the latent space. In this section, we attempt to characterize the nature of spurious fixed points more concretely.

\subsection{Rival Attractor}

\begin{figure}[h]
\vskip 0.2in
\begin{center}
\centerline{\includegraphics[width=\columnwidth]{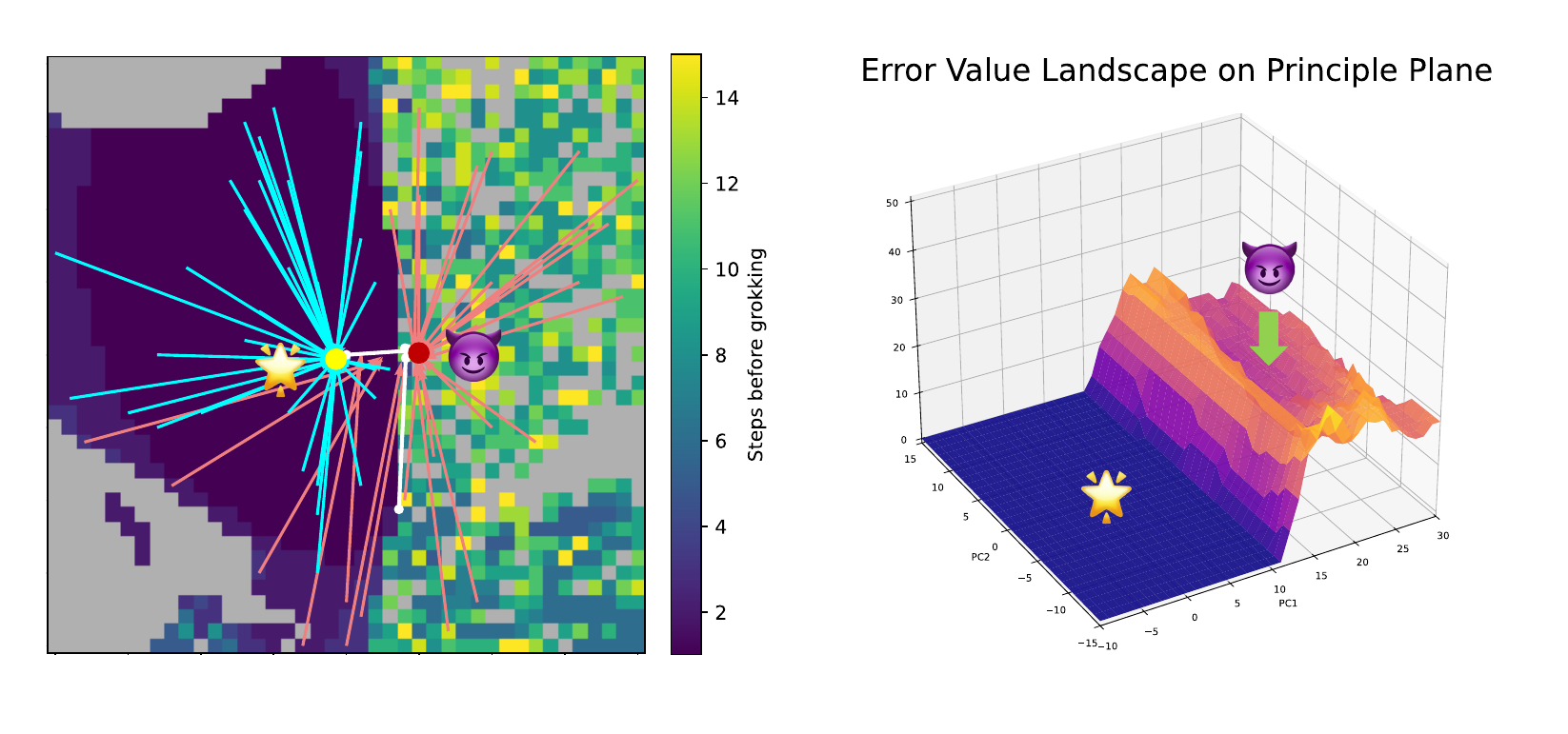}}
\caption{An example exhibiting rival attractors. \textbf{Left}: Arrows show the first update to the latent state at the starting point. Darker color in the background means that the model reaches the correct solution faster starting from there, while gray means a failure. There appear to be two attractors (yellow: correct; red: misleading), the interaction between which create a clear boundary of two regions. \textbf{Right}: The landscape of $\mathcal{E}$ value on the PCA plane. The ridge shows that this metric is non-monotonic (has an ``energy barrier") between the rival attractors.}
\label{fake_fp_figure}
\end{center}
\end{figure}

In \cref{fake_fp_figure}, both the true and spurious fixed points for a sample from the \textbf{nontrivial success} class are shown. The model first finds the false one, lingers for several segments before vaulting towards the true one. Both fixed points are indeed attractive points, in the sense that when the latent state is initialized close to either of them, it quickly gets updated to it.

The attractive effect of these two points competes with each other, creating a clear line of separation in the middle. The latent state is consistently attracted to the fixed point closer to it. Whenever $z^0$ is initialized closer to the true attractor, the model converges to it with the correct output in 1 or 2 steps. However, when initialized closer to the false one, it gets trapped there for large and varying numbers of segments, or even forever. Such phenomena are ubiquitous for the non-trivial success mode discussed in \cref{four_modes}.

\subsection{Plausible Local Minima}

In nonconvex optimization, gradient methods are susceptible to being trapped by local minima. This bears a strong resemblance to spurious fixed points being malignant attractors.

We hypothesize that the segment dynamics of HRM are implicitly doing minimization on some label-agnostic\footnote{If we are to compare inference-time updates to optimizing some function, this function has to be independent of the ground truth, since the model has no access to the label. For example, it would make no sense to say that ``HRM is performing gradient descent w.r.t. the loss function''.} energy function, measuring ``how good the current output is'' by counting conflicts. Here is one possible definition: for an output sudoku $\hat{y}$, we count the number $\operatorname{count}(d,u)$ of each token $d$ in the rows $r(\hat{y})$, columns $c(\hat{y})$ and boxes $b(\hat{y})$. If any of these counts exceeds $1$, we penalize this violation of sudoku rule by $\operatorname{count}(d,u)-1$. Formally, we define an error metric as:
\begin{equation}
    \mathcal{E}(\hat{y})=
    \sum_{u\in\{r(\hat{y}), c(\hat{y}), b(\hat{y})\}}\;
    \sum_{d=1}^{9}
    \operatorname{ReLU}\bigl(\operatorname{count}(d,u)-1\bigr)
\end{equation}
Obviously $\mathcal{E}=0$ if and only if $\hat{y}$ is a legal sudoku.

This metric helps partially explain the reason HRM lacks the incentive to escape spurious fixed points automatically. By evaluating $\mathcal{E}(f_O(z;\theta_O))$ with $z$ sampled around the rival attractors on the PCA plane, we find that the misleading attractor does appear to be a shallow local minimum of this metric (\cref{fake_fp_figure}). In particular, along the segment from the spurious fixed point to its true counterpart, $\mathcal{E}$ first slightly increases before dropping to $0$.

We point out that it would be premature to say that HRM actually implements an optimization algorithm on this quantity. 
Nevertheless, this opens up a new perspective for understanding HRMs, worthy of investigation in future work.

\section{Conclusion and Discussion}

The goal of this paper is to develop a mechanistic account of how hierarchical reasoning models ``reason''. Starting with the examination of fixed point property, we detect a violation of this fundamental assumption, which limits performance and causes several peculiar behaviors.

Fixed points are important not only because they ground the architecture of HRM mathematically, but also in that they are directly related to the correctness of HRM output. Via a series of qualitative experiments to understand HRM's seek for fixed points, we conclude that multiple fixed points exist in the latent space, but some correspond to false solutions. HRM ``guesses'' the solution by sticking to the first fixed point it encounters, which usually is the one closest to its initialization. We also develop a heuristic discrimination between true and false fixed points.

Utilizing the acquired understanding, we fix existing flaws of HRM, develop several simple but effective techniques to exploit its potential, and achieve significant performance enhancement with Augmented HRM.

Our analysis is primarily based on the recursive mechanics of HRM. Thus, while our experiments focus on HRM, the lens naturally extends to most recursive models. We conjecture that the qualitative taxonomy we provide will serve as a common vocabulary for the emerging class of recursive reasoners.

\section*{Impact Statement}

This paper presents work whose goal is to advance the field of machine learning. There are many potential societal consequences of our work, none of which we feel must be specifically highlighted here.



\bibliography{hrm}
\bibliographystyle{icml2026}

\newpage
\appendix
\onecolumn

\section{Related Work}

\textbf{Latent-space and Recursive Reasoning Models}
Rather than generating explicit chain-of-thought tokens, latent-reasoning models embed the intermediate computation in a continuous hidden trajectory. Coconut~\cite{hao2025traininglargelanguagemodels} chains hidden states to enable latent space planning; Heima~\cite{shen2025efficientreasoninghiddenthinking} compresses each step into one “thinking token”, cutting output length while maintaining precision. These models, featuring recursive latent-state updates, are emerging as a new paradigm for modeling reasoning~\cite{jolicoeurmartineau2025morerecursivereasoningtiny,zhang2025recursivelanguagemodels, guan2025rstarmathsmallllmsmaster}. Recursion decouples reasoning depth from parameter count, yielding sizable savings in memory and compute~\cite{dehghani2019universaltransformers}. HRM~\cite{wang2025hierarchicalreasoningmodel} -- one of the latest entries in this line -- introduces hierarchical modules trained with deep supervision to achieve reasoning depth scaling, and is the main subject of our work.

\textbf{Hierarchical Reasoning Model and Variants}
HRM is proposed by ~\citet{wang2025hierarchicalreasoningmodel}, achieving 55\% accuracy on \textit{Sudoku-Extreme}, surpassing latest LLM-based reasoners. ~\citet{ge2025hierarchicalreasoningmodelsperspectives} showed via ablation study that the hierarchical architecture does not significantly contribute to overall performance. They verified the fixed point property on \textit{Sudoku-Extreme}, but did not notice the violation on other samples. CGAR~\cite{qasim2025acceleratingtrainingspeedtiny} refines the HRM architecture and speeds up training. Another variant of HRM named Tiny Recursive Model~\cite{jolicoeurmartineau2025morerecursivereasoningtiny} emphasizes the role of recursive outer loop, achieved 87.4\% accuracy, at the cost of non-scalable reasoning depth. Our work achieves even better performance without altering the HRM architecture.

\textbf{Nature of Reasoning} Despite the performance boost achieved by both CoT and latent-space reasoning, there remain debates on what type of computation truly counts as `reasoning'~\cite{xu2025formal}. Various methods have been developed to gain a mechanistic understanding of CoT~\cite{bogdan2025thoughtanchorsllmreasoning,yao2025unveilingmechanismsexplicitcot, an2025dont} and of latent reasoning~\cite{zhang2025latenttokensthinkcausal}, mostly through the features of reasoning traces. Such features include monosemantic features extracted by sparse autoencoding~\cite{chen2025doeschainthoughtthink, theodorus2025finding} and geometric properties of the trace itself~\cite{zhou2025geometryreasoningflowinglogics, vilas2025tracingtraceslatenttemporal, du2025latentthinkingoptimizationlatent}.

\section{Results on Other Datasets}

The set of tools that we used for analyzing reasoning patterns is mostly independent of the specific task. Thus, most results are easily transferable to other datasets. In the original paper of~\citet{wang2025hierarchicalreasoningmodel}, besides \textit{Sudoku-Extreme}, \textit{Maze-Hard} data set is used to probe reasoning capabilities. This task involves finding the optimal path in a $30\times 30$ maze, with exactly one starting and one terminal point specified.

\begin{figure}[ht]
\vskip 0.2in
\begin{center}
\centerline{\includegraphics[width=0.8\columnwidth]{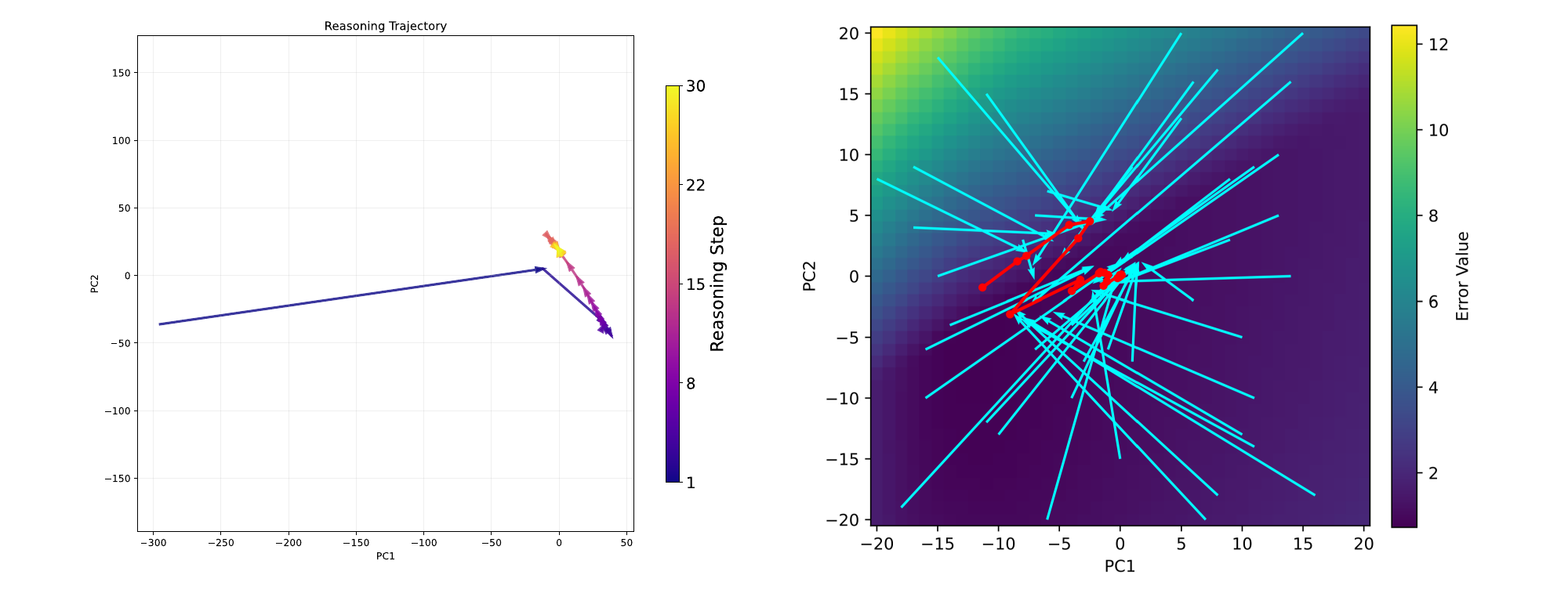}}
\caption{Reasoning trajectories for \textit{Maze-Hard} dataset. \textbf{Left}: The final-stage instability during inference, due to the violation of fixed-point assumption, is still present in most failure cases. \textbf{Right}: The multiple spurious attractors, despite being rarer, still exist; their rival effect on the reasoning trajectory is also similar.}
\label{maze_fig}
\end{center}
\end{figure}

The augmentation techniques devised in \cref{perturb} have natural counterparts for \textit{Maze-Hard}. Model bootstrapping is nearly identical (10 checkpoints from the second half of training, with interval of 1000 steps). We implement the token relabeling technique by swapping the starting and terminal points of a maze puzzle, creating an equivalent variant.
The evaluation results are shown in \cref{maze_results}.

\begin{table}[t]
\caption{Evaluation results of augmentation methods proposed in \cref{perturb}, on the \textit{Maze-Hard} dataset. Model bootstrapping and relabeling both improve accuracy as for \textit{Sudoku-Extreme}.}
\label{maze_results}
\begin{center}
\begin{small}
\begin{sc}
\begin{tabular}{lcccr}
\toprule
\textbf{Method} & \textbf{Exact Accuracy} \\
\midrule
Baseline & 74.5\% \\
+Bootstrap & 92.1\% \\
+Relabel & 76.7\% \\
+All (\textbf{Augmented HRM}) & \textbf{93.3\%} \\
\bottomrule
\end{tabular}
\end{sc}
\end{small}
\end{center}
\vskip -0.1in
\end{table}

The HRM reasoning trace of the maze task proves to be simpler than that of \textit{Sudoku-Extreme}. Successful cases typically have very short reasoning trajectories -- the solution is nearly always found in very early segments. In other words, the non-trivial success case in \cref{four_modes} is hardly present. Meanwhile, fixed point violation (\cref{violation}) is the dominant factor of failures. This might be because maze puzzles are intrinsically ``easier'' than sudoku puzzles, the latter involving more combinatorial structures that neural networks struggle to model~\cite{sudoku2018}.

Although spurious attractors do not constitute (\cref{misleading_attractors}) the main cause of failure in maze puzzles, we still identify their existence in some rarer cases. Their misleading effect on reasoning trajectories is evident as well (see \cref{maze_fig}).

\section{Implementation Details of HRM}
\label{simplification}

To facilitate exposition, the main paper condenses one full reasoning cycle into the single recurrence  
\[
z^{i+1}=\mathcal{F}(z^{i},\tilde{x};\theta).
\]  
This appendix exposes the sub-segment structure that was abstracted away, ensuring full reproducibility.  
Notation is inherited from the main text; additional sub-scripts $L$ (low-level) and $H$ (high-level) distinguish the two internal modules.

\subsection{Sub-segment Unfolding}
Each segment is implemented as $N\!\times\! T$ elementary time-steps (default $N\!=\!2$, $T\!=\!2$).  
Two latent trajectories are maintained:
\[
z_{L}^{t}\in\mathbb{R}^{d},\quad z_{H}^{k}\in\mathbb{R}^{d},\qquad
t=1{\dots}NT,\;k=0{\dots}N.
\]
The low-level module updates at every step; the high-level module updates once every $T$ steps, yielding a \emph{hierarchical convergence} schedule:
\begin{align*}
\intertext{\bf cycle $k=0$}
t=1{\dots}T\quad& z_{L}^{t}=f_{L}(z_{L}^{t-1},z_{H}^{0},\tilde{x};\theta_{L})\\
t=T\quad& z_{H}^{1}=f_{H}(z_{H}^{0},z_{L}^{T};\theta_{H})\\[2pt]
\intertext{\bf cycle $k=1$}
t=T\!+\!1{\dots}2T\quad& z_{L}^{t}=f_{L}(z_{L}^{t-1},z_{H}^{1},\tilde{x};\theta_{L})\\
t=2T\quad& z_{H}^{2}=f_{H}(z_{H}^{1},z_{L}^{2T};\theta_{H})\\
&\;\;\vdots\\
\intertext{\bf final output}
t=NT\quad& z^{i+1}\triangleq z_{H}^{N}.
\end{align*}
Both $f_{L}$ and $f_{H}$ are 4-layer transformer blocks ($d\!=\!512$, $8$ heads, Post-Norm with RMSNorm, with RoPE position encoding).

$z_{L}^{0}$ is always initialized as a deterministic tensor (trainable) inside a segment;  
$z_{H}^{0}$ is inherited from the previous segment's $z_{H}^{N}$ and detached to block gradient flow.  
This hard reset is claimed to force the low-level module to re-converge every segment, preventing premature saturation \cite{wang2025hierarchicalreasoningmodel}.

\subsection{ACT Implementation}
After obtaining $z_{H}^{N}$, a linear Q-head produces scalars
\[
Q_{\text{halt}},\;Q_{\text{continue}}=\mathrm{Linear}(z_{H}^{N})\,\in\mathbb{R}^{2}.
\]
A greedy rule decides whether to halt; if continuation is chosen, $z^{i+1}=z_{H}^{N}$ is fed into the next segment without extra parameters.

\subsection{Mapping to the Main-text Notation}
The abstract operator $\mathcal{F}(\cdot)$ in \cref{forward_z} of the main paper realizes the entire $N\!\times\! T$ hierarchy described above;  
$z^{i}$ corresponds to $z_{H}^{0}$ entering the segment, and $z^{i+1}$ corresponds to $z_{H}^{N}$ exiting it. The operator $\mathcal{F}(\cdot)$ is applied $M$ (default $M=16$) times as the outer loop (described in \cref{forward_pass}).
All low-level dynamics are encapsulated inside $\mathcal{F}$, so the simplified description remains functionally identical to the original HRM.


\end{document}